\documentclass[lettersize,journal]{IEEEtran}
\usepackage{amsmath,amsfonts}
\usepackage{algorithmic}
\usepackage{algorithm}
\usepackage{array}
\usepackage[caption=false,font=normalsize,labelfont=sf,textfont=sf]{subfig}
\usepackage{textcomp}
\usepackage{stfloats}
\usepackage{url}
\usepackage{verbatim}
\usepackage{graphicx}
\usepackage{cite}
\usepackage{xcolor}
\usepackage{booktabs}
\usepackage{multirow} 

\begin{document}

\title{Towards Comprehensive Stage-wise Benchmarking of Large Language Models in Fact-Checking}

\author{Hongzhan~Lin,
        Zixin~Chen,
        Zhiqi~Shen,
        Ziyang~Luo,\\
        Zhen~Ye,
        Jing~Ma,
        Tat-Seng~Chua
        and Guandong~Xu
\IEEEcompsocitemizethanks{\IEEEcompsocthanksitem Hongzhan Lin, Zixin Chen, Jing Ma are with the Department
of Computer Science, Hong Kong Baptist University. (\textit{E-mail: cshzlin@comp.hkbu.edu.hk})\protect
\IEEEcompsocthanksitem Zhiqi Shen, Ziyang Luo are with Salesforce AI Research.\protect
\IEEEcompsocthanksitem Zhen Ye is with The Hong Kong University of Science and Technology.\protect
\IEEEcompsocthanksitem Tat-Seng Chua is with National University of Singapore.\protect
\IEEEcompsocthanksitem Guandong Xu is with The Education University of Hong Kong.\protect}
}



\maketitle

\begin{abstract}
Large Language Models (LLMs) are increasingly deployed in real-world fact-checking systems, yet existing evaluations focus predominantly on claim verification and overlook the broader fact-checking workflow, including claim extraction and evidence retrieval. This narrow focus prevents current benchmarks from revealing systematic reasoning failures, factual blind spots, and robustness limitations of modern LLMs. To bridge this gap, we present FactArena, a fully automated arena-style evaluation framework that conducts comprehensive, stage-wise benchmarking of LLMs across the complete fact-checking pipeline. FactArena integrates three key components: (i) an LLM-driven fact-checking process that standardizes claim decomposition, evidence retrieval via tool-augmented interactions, and justification-based verdict prediction; (ii) an arena-styled judgment mechanism guided by consolidated reference guidelines to ensure unbiased and consistent pairwise comparisons across heterogeneous judge agents; and (iii) an arena-driven claim-evolution module that adaptively generates more challenging and semantically controlled claims to probe LLMs’ factual robustness beyond fixed seed data. Across 16 state-of-the-art LLMs spanning seven model families, FactArena produces stable and interpretable rankings. Our analyses further reveal significant discrepancies between static claim-verification accuracy and end-to-end fact-checking competence, highlighting the necessity of holistic evaluation. The proposed framework offers a scalable and trustworthy paradigm for diagnosing LLMs’ factual reasoning, guiding future model development, and advancing the reliable deployment of LLMs in safety-critical fact-checking applications.
\end{abstract}


\section{Introduction}

\IEEEPARstart{L}{arge} language models (LLMs) have significantly advanced the field of natural language processing (NLP), demonstrating remarkable improvements across numerous tasks~\cite{touvron2023llama, OpenAI2023GPT4TR}. Previous studies have particularly highlighted the capability of LLMs to perform different tasks in the fact-checking pipeline~\cite{pan2023fact, wang2023explainable, zhao2024pacar}. Despite these advancements, LLMs remain vulnerable to factual inaccuracies and are prone to errors in reasoning~\cite{lin2022teaching, bubeck2023sparks}. Mistakes arising from flawed memorized knowledge or inadequate factual reasoning, can undermine their reliability and effectiveness in fact-checking applications~\cite{elazar2021measuring, cao2021knowledgeable, lin2022amif}. Consequently, it is critical to systematically identify and understand the capabilities of LLMs in fact-checking to ensure their robustness and enhance their overall trustworthiness.

\begin{figure}[t!]
    \centering    
    \scalebox{0.5}{\includegraphics[width=\textwidth]{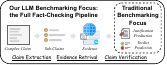} }
\vspace{-0.7cm}
    \caption{The comparison of different focuses between our proposed FactArena and traditional solutions in fact-checking evaluation. Different from the traditional solutions that only focus on claim verification, FactArena aims to scrutinize the full stages of the fact-checking pipeline. 
}
\vspace{-0.3cm}
    \label{fig:motivation}
\end{figure}

As illustrated in Figure~\ref{fig:motivation}, the fact-checking process~\cite{guo2022survey} typically consists of three critical stages: (i) claim extraction, where claims requiring verification are identified; (ii) evidence retrieval, where relevant sources supporting or refuting these claims are gathered; and (iii) claim verification, which involves assessing the veracity of the claims based on the retrieved evidence and providing justifications for the resulting verdict. Traditional approaches~\cite{hu2024large, lin2025fact} for evaluating LLM performance in fact-checking have primarily focused on the last stage: claim verification, while largely neglecting the equally critical earlier stages.

Benchmarking LLM performance comprehensively across all fact-checking stages is essential, not solely claim verification. LLMs are increasingly deployed throughout the entire fact-checking pipeline, especially when addressing complex real-world claims. Overlooking any stage, especially the 
two preliminary stages, could severely compromise the integrity and effectiveness of the overall fact-checking systems. Thus, a thorough and systematic assessment of LLM capabilities at every stage is indispensable. A comprehensive auditing can yield deeper insights into the model limitations, thereby significantly advancing the reliable deployment of LLMs in practical fact-checking scenarios. Designing such an evaluation is inherently challenging, as the different tasks of the fact-checking pipeline may involve open-ended judgments that lack clear-cut ground truth. A potential remedy lies in leveraging platforms such as Chatbot Arena~\cite{chiang2024chatbot}, which crowdsource annotations from diverse users and employ pairwise comparisons to yield more nuanced and objective assessments of LLM capabilities. However, this strategy requires considerable data preparation and annotation resources~\cite{zheng2023judging}, rendering it prohibitively costly when applied at scale.

To address these challenges, we propose a systematic, automated arena-style framework that evaluates LLMs across the entire fact-checking pipeline using unbiased collective judgments. The goal is to facilitate comprehensive and reliable assessments of LLMs’ fact-checking capacities. Our framework is designed around three key principles: 1) The framework scrutinizes all phases of the fact-checking process rather than limiting evaluation to claim verification, thereby capturing a holistic view of model capabilities. 2) The framework must integrate diverse evaluator perspectives to ensure fair assessments of LLM performance. 
To ensure reliability as well as minimizing subjectivity in a single judgment,  
we draw on the notion of collective intelligence~\cite{leimeister2010collective}, aligning judgments across heterogeneous evaluators by establishing consensus on factuality. 3) The framework automates the evaluation process to enable efficient and large-scale assessments. Inspired by agentic evaluation approaches~\cite{park2023generative, gu2024survey}, we modularize the auditing of LLMs, thereby reducing reliance on costly manual annotation.

In this work, we introduce a multi-agent evaluation framework, \textit{FactArena}, to systematically benchmark the fact-checking capabilities of LLMs across all stages of the pipeline in an arena-style fashion. Specifically, we first design an automatic fact-checking pipeline deployed with LLMs. Then, aligned with the fact-checking process, our framework is designed with stage-wise peer-battle arenas in three steps:
(i) assessing the ability of LLMs to decompose complex claims and plan verification tasks; (ii) evaluating their capacity to retrieve and integrate auxiliary evidence through tool-augmented interactions; and (iii) auditing the quality of their explanations and the soundness of their final verdicts. Besides, we incorporate an arena-driven claim evolution mechanism that generates progressively more challenging check-worthy claims based on model performance, thereby enabling more adaptive and model-centric evaluation. By unifying these stages under a scalable, automated evaluation paradigm, \textit{FactArena} offers a holistic and unbiased lens for understanding LLMs’ strengths and limitations. We believe this framework provides a critical step toward advancing rigorous, reliable, and trustworthy deployment of LLMs in real-world fact-checking applications.

Our contributions can be summarized as follows:
\begin{itemize}
    \item To the best of our knowledge, this is the first work to fully automate the trustworthy benchmarking of LLMs across the entire fact-checking pipeline. Our framework provides an analytical lens for assessing stage-wise factual reasoning under diverse, open-form evaluation scenarios.
    \item We propose \textbf{FactArena}, an arena-style, agent-based evaluation framework that enables fair and comprehensive evaluations that address the inherently open-ended and multi-stage nature of fact-checking.
    \item We introduce an arena-driven claim evolution module that adaptively generates increasingly challenging and unseen claims. This mechanism systematically probes the factual robustness and knowledge boundaries of LLMs, offering a dynamic complement to fixed test sets.
    \item Extensive experiments demonstrate that FactArena substantially improves the reliability and wholeness of LLM evaluation in the fact-checking workflow. Judgments from diverse agent evaluators exhibit strong consistency and align closely with human expert assessments, providing actionable insights for trustworthy LLM auditing.
\end{itemize}



\section{Related Work}
\paragraph{Fact-Checking Evaluation} Automated fact-checking has received growing attention in the NLP community as a strategy to combat misinformation and disinformation. Over the past few years, a variety of datasets have been introduced to support the development and evaluation of fact-checking systems. These include collections based on: human-curated claims from Wikipedia~\cite{thorne2018fever, sathe2020automated, schuster2021get}; fabricated claims from news outlets~\cite{tariq2016nelasso, buntain2017automatically, shu2020fakenewsnet, nakov2022clef}; rumorous claims on social media~\cite{ma2015detect, ma2017detect, lin2021rumor}; complex claims requiring multi-hop reasoning~\cite{jiang2020hover, aly2021feverous}; naturally occurring claims in specialized domains~\cite{gupta2021x, wadden2022scifact, lin2022detect, lin2023zero}; and, more recently, misinformation generated by LLMs~\cite{chen2024can}.
To assess the factual knowledge of LLMs, \cite{hu2024large} consolidated several representative datasets into a unified benchmark, aiming to reveal weaknesses in LLM-based fact verification. However, in addition to the inevitable risk of test set leakage, this static evaluation paradigm largely depends on expert-designed datasets and specialized tasks. Such reliance limits adaptability to emerging forms of misinformation, particularly LLM-generated content, and falls short in capturing the open-ended complexity of real-world applications.
Distinct from prior static accuracy evaluations, our work leverages the justifications produced by LLMs~\cite{atanasova2020generating, guo2022survey} to enable dynamic auditing. Rather than focusing solely on veracity prediction, we seek to actively elicit and analyze the limitations of LLMs in fact-checking, thereby offering a more adaptive and fine-grained understanding of their capabilities.


\paragraph{Multi-agent System} Deliberation among multiple agents can enhance factual accuracy and robustness through debate and consensus. To this end, multi-agent frameworks have been developed to address complex tasks through coordinated interactions. For instance, MetaGPT~\cite{hong2024metagpt}, ChatDev~\cite{qian2024chatdev}, and AgentVerse~\cite{chen2024agentverse} assign specialized roles and enable information sharing and cross-checking via natural language dialogue, often surpassing single-agent approaches~\cite{tran2025multi}. Role-playing paradigms such as CAMEL~\cite{li2023camel} further illustrate how inception prompting supports autonomous collaboration while reducing reliance on human supervision. Beyond task-specific applications, generative multi-agent environments~\cite{park2023generative, patil2024gorilla} demonstrate that agents equipped with memory and reflection can simulate complex social interactions and long-term behavioral dynamics.
While LLM-based multi-agent evaluation has been applied to audit fact-checking performance in the previous work~\cite{lin2025fact}, it largely overlooks the equally critical stages of claim extraction and evidence retrieval, primarily focusing on claim verification. This gap underscores the need for comprehensive evaluation~\cite{xu2024lvlm} across the full stage-wise fact-checking pipeline, particularly for assessing LLM capacities in claim extraction, evidence retrieval, and justification production, where no closed-form ground truth is available for direct benchmarking.

\begin{figure*}[t!]
    \centering    
    \includegraphics[width=\textwidth]{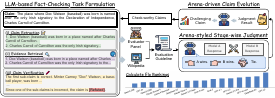} 
\vspace{-0.7cm}
    \caption{An overview of our proposed FactArena framework. We automatically conduct a comprehensive benchmarking process for large language models in complete fact-checking stages (e.g., claim extraction, evidence retrieval, and claim verification), distinct from previous traditional audits only focused on the claim verification stage, which is only part of the full fact-checking pipeline.
}
\vspace{-0.3cm}
    \label{fig:overview}
\end{figure*}

\section{Methodology}
\subsection{Overview}
\textbf{Problem Statement.} Given a complex claim $C$, fact-checking aims to predict the factuality and provide convincing justifications, to evaluate the claim as supportive (i.e., true) or refuted (i.e., false), based on evidence as auxiliary information. Our objective is to develop an agent-driven evaluation framework, which aims to conduct a comprehensive and unbiased arena-style evaluation that automatically assesses LLMs’ stage-wise abilities to fact-check complex claims.

To thoroughly scrutinize the fact-checking capacities of LLMs in different fact-checking stages, the FactArena framework consists of the following three major components: 1) Construct the completed fact-checking process driven by LLMs (\S\ref{sec:lfp}); 2) Conduct peer-battle and automatic judgment by each fact-checking stage in an arena-like manner (\S\ref{sec:ajs}); 3) Generate progressively challenging check-worthy claims based on model performance to further assess the model’s capabilities (\S\ref{sec:ace}). An overview of our FactArena framework is shown in Figure~\ref{fig:overview}.

\subsection{LLM-based Fact-Checking Task Formulation} \label{sec:lfp}
To comprehensively benchmark the fact-checking ability of LLMs, we need to construct a completed fact-checking pipeline by deploying the target model as the core controller. The pipeline consists of three sequential stages: 1) Claim Extraction; 2) Evidence Retrieval; and 3) Justification Production \& Verdict Prediction (i.e., Claim Verification).

\paragraph{Claim Extraction} It refers to decomposing a complex claim into verifiable sub-claims, which then serve as the basis for planning the subsequent fact-checking process. This formulation aligns claim extraction with task decomposition, ensuring that each sub-claim can be systematically verified through evidence retrieval and justification generation. Specifically, for the complex claim $C$ in our designed fact-checking pipeline, we first employ the target LLM to decompose it as a set of sub-claims: 
\begin{equation}
    \{c_1, c_2, ..., c_k\} \leftarrow \text{LLM}(C),
\end{equation} where $k$ is the number of decomposed sub-claims. 

\paragraph{Evidence Retrieval} It aims to find and summarize factual information beyond the claim from external knowledge sources to indicate veracity. Specifically, the target LLM would integrate the external tool (i.e., Google Search) to retrieve potentially relevant information. 
The claim $C$ 
is used as the query, and we collect the web information (i.e., the title and snippet) of the top-ranked result as the initial contextual source, as follows:
\textcolor{black}{\begin{equation}
    w \leftarrow \text{Search}(C),
\end{equation} where $w$ is the retrieved web information for the claim $C$}, $\text{Search}(\cdot)$ denotes the function of Google Search. Subsequently, the target LLM processes the claim together with its decomposed sub-claims and the retrieved web information. Based on this input, the model extracts and summarizes the key factual information that can serve as evidence for claim verification:
\begin{equation}
    e \leftarrow \text{LLM}(C, \{c_1, ..., c_k\}, w),
\end{equation} where $e$ means the output evidence. This step ensures that the evaluation leverages both the target model’s reasoning over decomposed sub-claims and its ability to identify salient facts from external sources, thereby constructing evidence to support or refute the original claim.

\paragraph{Justification Production \& Verdict Prediction} It targets yielding a reasonable justification for fact-checking the original claim, ultimately leading to the predicted veracity label. Specifically, the target LLM executes the claim verification stage by processing the original claim, the sub-claims, and the retrieved evidence obtained in previous fact-checking stages, formulated as:
\begin{equation}
    \{r, y\} \leftarrow \text{LLM}(C, \{c_1,...,c_k\}, e),
\end{equation} where $r$ and $y$ denote the generated justification and the predicted verdict, respectively, forming the final verification output.

\begin{figure*}[t!]
    \centering    
    \includegraphics[width=\textwidth]{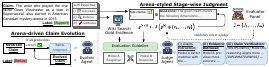} 
\vspace{-0.7cm}
    \caption{An illustration of arena-styled stage-wise judgment and arena-driven claim evolution.
}
\vspace{-0.3cm}
    \label{fig:method}
\end{figure*}

\subsection{Arena-styled Stage-wise Judgment} \label{sec:ajs}
By deploying the target LLM to strictly follow the complete fact-checking workflow, we have first provided a transparent platform for evaluating its capabilities at each stage of the process. A key challenge is to establish a comprehensive benchmark system that assesses model performance across all fact-checking stages, accommodates their open-form outputs, and seamlessly integrates stage-wise abilities into an overall evaluation of fact-checking competence. Therefore, we propose the arena-styled stage-wise judgment approach: We initiate peer battles between two anonymous models, following the Chatbot Arena style, where two models are randomly chosen from the model pool, to generate the stage-wise response (e.g., sub-claims, evidence, or verification output).
Our goal is to fairly compare the overall quality of stage-wise responses, thereby providing a transparent basis for determining which target model demonstrates superior performance across the full fact-checking pipeline.

Specifically, on a set $\mathcal{C}$ of the complex claim data, we evaluate a group of target LLMs $\{\mathcal{T}_1,...,\mathcal{T}_n\}$, with an evaluator panel $\mathbb{J}$ consisting of judge agents as follows:
\begin{equation}
\mathcal{R} \leftarrow \text{FactArena}(\mathcal{T}_1,...,\mathcal{T}_n|\mathbb{J},\mathcal{C}),
\end{equation}
where the evaluator panel $\mathbb{J} = \{{J_1, \dots, J_m}\}$ consists of $m$ judge agents, and $\mathcal{R}$ denotes the stage-wise evaluation outcomes in the form of a ranking that reflects the relative capabilities of the $n$ ($n>m$) target LLMs in fact-checking complex claims.

Any competitive evaluation requires clear reference guidelines to minimize biases arising from differences among committee members~\cite{zhang2021identifying, gehrmann2022gemv2}. To this end, we first establish evaluation guidelines for the open-form responses produced in the three stages of fact-checking: claim extraction, evidence retrieval, and claim verification.

For claim extraction, fair evaluation of decomposed sub-claims requires evaluators to reach a holistic consensus on what constitutes an appropriate breakdown. Thus, the judge agents begin by consolidating the diverse sub-claims generated by different target models, synthesizing their strengths to construct the most impartial guideline. This unified reference provides a consistent basis for judges to assess model performance at this stage. For each claim $C$ with sub-claim sets from $n$ target models, an evaluator panel $\mathbb{J}$ integrates these different ways of claim decomposition to refine the evaluation guideline through multiple rounds.

To mitigate bias, the panel is diversified by selecting strong judge agents from different model families, and judges are not allowed to evaluate their own responses. In the current $r$-th round, a judge $J$ examines the current guideline $g^{(r)}$ and a randomly sampled set of sub-claims, then proposes an updated version:
\begin{equation}
    g^{(r+1)} \leftarrow J( g^{(r)} \oplus \{c_1, ..., c_k\}).
\end{equation} Judges and sub-claim sets are sampled in rotation until all candidate sub-claim sets are incorporated, ensuring fairness and coverage. The process initializes with a randomly selected judge response and terminates once all viewpoints are integrated into a consolidated guideline. 

For evidence retrieval, to define a reliable reference information source, judge agents search Wikipedia using the entities mentioned in the complex claim, with the gold evidence from Wiki knowledge as the reference guideline for judgment. Judge agents then evaluate the evidence generated by target models against this guideline. In this way, model outputs are assessed by their ability to capture the essential factual content needed to support or refute the claim. Note that here the evidence guideline is not used as the golden label, but instead as the factual basis to further mitigate the potential factual errors inherent in the judge committee.

For claim verification, verdict prediction can be directly evaluated as a classification task; however, justification production requires additional guidelines to support arena-style evaluation. Justifications focus on whether the reasoning throughout the fact-checking pipeline is coherent, e.g., whether the model effectively leverages its extracted key information and provides a sound explanation for its final verdict. Therefore, rather than attempting to consolidate all model-generated justifications into a single comprehensive reference, inspired by previous automated judging benchmarks~\cite{zheng2023judging, li2024crowdsourced}, we establish evaluation guidelines~\cite{wang2024explainable} from the following perspectives:
\begin{itemize}
    \item \textit{Helpfulness}: The justification should align with the final verdict of the claim, ensuring consistency and avoiding misleading or contradictory reasoning.
    \item \textit{Informativeness}: The justification should identify and include all salient facts and key points from the evidence that are essential for verifying the claim.
    \item \textit{Soundness}: The justification should demonstrate logical validity and coherence, with reasoning that is well-structured and adequately supported by evidence.
    \item \textit{Readability}: The justification should be clearly articulated, grammatically correct, and easy to follow, allowing evaluators to understand the reasoning without difficulty.
\end{itemize}

Based on these guidelines, judge agents from the evaluator panel are employed in a manner similar to the LLM-as-a-Judge paradigm~\cite{zheng2023judging}, to determine and explain which target LLM produces the superior performance for the full fact-checking pipeline by scrutinizing their stage-wise outputs. 

After collecting pairwise judgments from the evaluator agents, we compute model rankings using the Elo rating system~\cite{elo1966uscf}, which iteratively updates scores based on head-to-head outcomes. The Elo framework estimates the probability that an LLM $a$ will outperform another LLM $b$, given their current ratings $R_a$ and $R_b$, where $a, b \in \mathbb{N}$. For each comparison, we define a binary outcome variable $Y_{ab}$, which equals $1$ if model $a$ is judged the winner and $0$ otherwise. The predicted probability is then expressed as:
\begin{equation}
    P(Y_{ab} = 1) = \frac{1}{1 + 10^{(R_b - R_a)/\alpha}},
\end{equation}
where $\alpha = 400$ serves as the scaling constant in the Elo formula.
In the standard Elo algorithm, model ratings are updated according to: $R'_a = R_a + K \cdot \left(S(a, b) - P(Y_{ab} = 1)\right)$, where $K$ is a scaling constant and $S(a, b)$ represents the observed outcome for LLM $a$ in its comparison with LLM $b$ (taking values 1 for a win, 0.5 for a draw, and 0 for a loss). Although the Elo framework effectively models pairwise win probabilities, it performs sequential updates that make the resulting scores sensitive to the order of match-ups. To mitigate this limitation and achieve more stable rankings, we additionally employ the Bradley–Terry model, which estimates relative strengths without relying on update order.

The Bradley–Terry algorithm~\cite{bradley1952rank} provides a probabilistic framework for ranking, extending Elo-based evaluations by treating pairwise outcomes as logistic comparisons and estimating model strengths via maximum likelihood. For a set of $n$ models with observed pairwise results, let $W_{ab}$ denote the number of times LLM $a$ defeats LLM $b$. The overall log-likelihood of the comparisons is then formulated as:
\begin{equation}
    \mathcal{L}(\mathbb{R}) = \sum_{a \ne b} W_{ab} \cdot \log P(Y_{ab} = 1),
\end{equation}
where $\mathbb{R} = \{R_1, R_2, \dots, R_n\}$ are the model ratings.
Because the Bradley–Terry algorithm does not natively account for ties, we handle tie votes by splitting them evenly: each tie is recorded as half a win for both models, incrementing both $W_{ab}$ and $W_{ba}$ by $0.5$. This adjustment ensures a fair and balanced estimation of model rankings across all pairwise comparisons. Finally, by sorting the estimated model ratings $\mathbb{R}$, we obtain the ranking $\mathcal{R}$ of the target LLMs, which serves as the ultimate evaluation outcome for the fact-checking pipeline.

\subsection{Arena-driven Claim Evolution} \label{sec:ace}

Analogous to the use of hard prompts in Chatbot Arena~\cite{li2024crowdsourced}, introducing more challenging claims can further advance the evaluation of LLMs’ fact-checking abilities. A central challenge, however, lies in identifying the specific areas where a target model underperforms. While existing fact-checking datasets provide an intuitive yet superficial assessment of model performance, we argue that they are insufficient for uncovering the deeper limitations and knowledge boundaries of LLMs. This limitation stems from the inherent constraints of fixed test sets, which may also introduce risks of test leakage. 

To craft diverse and harder claims when the target models all predict the correct verdict, we propose iteratively evolving for a more comprehensive fact-checking evaluation by drawing insights from the model behaviors of the arena-like auditing records as environmental feedback. Specifically, if all the target models predict the correct factuality of the claim $C$, we first employ the evolver agent to reverse the semantics of the original claim $C$, thereby generating a contrastive claim $\hat{C}$ with the opposite verdict. If the factuality of the contrastive claim is still correctly predicted by all target models, we then initiate an iterative targeted evolution process driven by the arena battle judgments, generating semantically equivalent but increasingly challenging variations of the claim. 

Note that each evolved claim is evaluated by judge agents against the target LLMs, with the results stored in a judgment record pool. By iteratively evolving claims in this manner, we can identify test instances where models perform poorly under specific fact-checking scenarios, thereby yielding comprehensive insights into their fact-checking capacities across diverse complex claims.

\begin{table}[t] \small
\centering
\caption{The statistics of claims and battle count in different stages of FactArena framework, where \textit{OC} means \textit{Original Claim} and \textit{EC} means \textit{Evolved Claim}.}
\vspace{-0.3cm}
\label{tab:statistics}
\renewcommand{\arraystretch}{1.1}
\begin{tabular}{l|c|ccc}
\toprule 
\multirow{2}{*}{\textit{Models}} & {\textit{Claim}} & {\textit{Battle}} & {\textit{Battle}} & {\textit{Battle}}\\
 &  & \textit{(OC)} & \textit{(EC)}& \textit{(Total)}\\
\midrule
\textbf{Claude Opus 4} & 213 & 1148 & 358 & 1506 \\
\textbf{Claude Sonnet 4} & 216 & 1006 & 299 & 1305 \\
\textbf{DeepSeek-R1} & 214 & 1157 & 427 & 1584 \\
\textbf{DeepSeek-V3} & 208 & 1136 & 414 & 1550 \\
\textbf{GPT-4.1} & 241 & 1326 & 466 & 1792 \\
\textbf{GPT-4.5} & 240 & 1315 & 469 & 1784 \\
\textbf{GPT-4o} & 210 & 1159 & 434 & 1593 \\
\textbf{GPT-o3} & 224 & 1269 & 357 & 1626 \\
\textbf{GPT-o4 mini} & 237 & 1289 & 450 & 1739 \\
\textbf{Gemini 2.5 Flash} & 245 & 1325 & 476 & 1801 \\
\textbf{Gemini 2.5 Pro} & 231 & 1259 & 457 & 1716 \\
\textbf{Grok-3} & 215 & 1196 & 361 & 1557 \\
\textbf{Grok-3 mini} & 219 & 1180 & 459 & 1639 \\
\textbf{Llama 4 Maverick} & 221 & 1221 & 416 & 1637 \\
\textbf{Qwen3 (235B)} & 234 & 1299 & 438 & 1737 \\
\textbf{Qwen3 (32B)} & 200 & 1137 & 259 & 1396 \\
\bottomrule
\end{tabular}
\end{table}

\begin{table*}[t] \large
\centering
\caption{Model performance rankings across multiple evaluation dimensions. \textbf{Battle Count} indicates the number of pairwise comparisons each model participated in. \textbf{Acc. (\%)} denotes the accuracy percentage. The ranking is sorted based on \textbf{Overall Judge}.}
\vspace{-0.3cm}
\label{tab:main_results}
\renewcommand{\arraystretch}{1.1} 
\resizebox{\textwidth}{!}{
\begin{tabular}{l|c|cccccc|cc}
\toprule 
\multirow{2}{*}{{Target mLLMs}} & \textit{Battle} & \textit{Claim} & \textit{Evidence} & \textit{Justification} & \textit{Justification} & \textit{Justification} & \textit{Justification} & \textit{Overall} & \textit{Acc.} \\
 & \textit{Count} & \textit{Extraction} & \textit{Retrieval} & \textit{-Helpfulness} & \textit{-Informativeness} & \textit{-Soundness} & \textit{-Readability} & \textit{Judge} & (\%) \\
\midrule
\textbf{GPT-o3 (-)} & 1626 & \textbf{1188.11} & \textbf{1351.95} & \textbf{1294.67} & \textbf{1331.77} & \textbf{1292.55} & \textbf{1268.83} & \textbf{1320.14} & \underline{65.02} \\
\textbf{DeepSeek-R1 (671B)} & 1584 & \underline{1096.22} & \underline{1129.69} & \underline{1141.88} & \underline{1160.02} & \underline{1131.83} & \underline{1117.53} & \underline{1127.77} & 55.14 \\
\textbf{GPT-o4 mini (-)} & 1739 & 1081.00 & 1129.67 & 1094.46 & 1101.49 & 1097.33 & 1097.51 & 1108.47 & 64.14 \\
\textbf{Gemini 2.5 Pro (-)} & 1716 & 1046.19 & 1066.15 & 1085.29 & 1078.02 & 1081.98 & 1077.44 & 1084.19 & \textbf{66.52} \\
\textbf{GPT-4.5 (-)} & 1784 & 1050.66 & 1041.89 & 1052.08 & 1039.95 & 1053.24 & 1041.03 & 1047.15 & 59.41 \\
\textbf{GPT-4.1 (-)} & 1792 & 1018.24 & 1033.21 & 1044.00 & 1032.77 & 1044.03 & 1029.11 & 1041.71 & 64.32 \\
\textbf{Grok-3 (-)} & 1557 & 995.75 & 1007.77 & 1027.00 & 1026.11 & 1026.05 & 1019.52 & 1022.09 & 55.14 \\
\textbf{Grok-3 mini (-)} & 1639 & 979.61 & 960.10 & 965.29 & 966.40 & 958.97 & 974.06 & 960.74 & 52.75 \\
\textbf{Gemini 2.5 Flash (-)} & 1801 & 969.48 & 955.35 & 962.62 & 945.27 & 956.76 & 980.50 & 955.60 & 55.51 \\
\textbf{Claude Opus 4 (-)} & 1506 & 973.20 & 949.44 & 953.03 & 961.30 & 958.32 & 933.39 & 949.27 & 44.34 \\
\textbf{DeepSeek-V3 (671B)} & 1550 & 968.84 & 924.27 & 937.19 & 923.45 & 938.83 & 951.61 & 934.38 & 55.77 \\
\textbf{Llama 4 Maverick (400B)} & 1637 & 919.58 & 916.68 & 922.33 & 926.36 & 916.31 & 909.13 & 919.57 & 60.91 \\
\textbf{Claude Sonnet 4 (-)} & 1305 & 940.14 & 908.29 & 905.29 & 930.52 & 919.29 & 892.76 & 905.28 & 34.72 \\
\textbf{Qwen3 (235B)} & 1737 & 936.05 & 892.18 & 893.44 & 882.68 & 891.82 & 922.65 & 890.89 & 55.13 \\
\textbf{GPT-4o (-)} & 1593 & 937.60 & 888.55 & 888.45 & 872.55 & 891.96 & 913.03 & 890.75 & 50.95 \\
\textbf{Qwen3 (32B)} & 1396 & 899.33 & 844.80 & 832.96 & 821.33 & 840.73 & 871.89 & 841.99 & 51.76 \\
\bottomrule
\end{tabular}}
\end{table*}

\section{Experiments}\label{sec:exp}
\subsection{Experimental Settings} \label{sec:exp_set}
\paragraph{Dataset}
We conduct experiments on complex claims selected from two publicly available fact-checking datasets, HOVER~\cite{jiang2020hover} and FEVEROUS~\cite{aly2021feverous}, that require multi-hop reasoning beyond simple factual matching to verify in fact-checking~\cite{pan2023fact}. The detailed data statistics are shown in Table~\ref{tab:statistics}.

\paragraph{Models}
We comprehensively evaluate a total of 16 LLMs across 7 distinct model families, covering a broad range of architectures and scales. Specifically, the models include: 1) Claude 4 Opus, Claude 4 Sonnet; 2) DeepSeek-R1, DeepSeek-V3; 3) GPT-4.5, GPT-4.1, GPT-o4 mini, GPT-o3, GPT-4o; 4) Gemini 2.5 Flash, Gemini 2.5 Pro; 5) Grok-3, Grok-3 mini; 6) Llama 4 Maverick; 7) Qwen3 (235B, 32B), 
from which the strong judge agents, DeepSeek-V3, GPT-o4 mini, Gemini 2.5 Flash, and Qwen3 (235B), are selected after balancing their performance and computational cost for the evaluator panel.

\subsection{Implementation Details} \label{sec:impl_details}

\paragraph{Model Details}
In FactArena, we conduct evaluation on a total of 16 LLMs across 7 distinct model families with the following representative versions:
1) Claude Opus 4: claude-opus-4-20250514;
2) Claude Sonnet 4: claude-sonnet-4-20250514;
3) DeepSeek-R1: deepseek-r1;
4) DeepSeek-V3: deepseek-v3;
5) Gemini 2.5 Flash: gemini-2.5-flash;
6) Gemini 2.5 Pro: gemini-2.5-pro-preview-06-05;
7) GPT-4.1: gpt-4.1-2025-04-14;
8) GPT-4.5: gpt-4.5-preview-2025-02-27;
9) GPT-4o: gpt4o;
10) GPT-o3: o3-2025-04-16;
11) GPT-o4 mini: o4-mini-2025-04-16;
12) Grok-3: grok-3;
13) Grok-3 mini: x-ai/grok-3-mini-beta;
14) Llama 4 Maverick: llama-4-maverick-17b-128e-instruct;
15) Qwen3 (235B): qwen3-235b-a22b;
16) Qwen3 (32B): qwen3-32b;
In arena-driven claim evolution, we utilize GPT-o4 mini, one of the dominant mLLMs, as the agent controller to reverse the semantics and create harder claims. We set the temperature parameter of all the target models and judges as 0.0 to guarantee reproducibility of our evaluation results as much as possible.

\paragraph{Data Statistics}
We randomly selected a total of 400 complex claims, each 200 from the datasets mentioned in ~\S\ref{sec:exp_set}. For each claim we sampled 8 target model responses per task for pairwise comparisons to ensure diversity of samples in model comparisons as well as to maintain a controllable number of total battles, following the combinatorial coverage theory~\cite{kuhn2013combinatorial}. In the selected 400 claims, 85 are correctly predicted by all the selected target models, which are then semantically reversed and evolved. The final arena-styled judgment results in about 13,000 valid judgments, with each target LLM participating in approximately 1,600 comparisons, roughly 104 times for each model pair on average. Compared results are averaged over three random 3 runs. The cost of API for evaluating one target model is about 15 dollars and 3 hours.

\subsection{Main Results}
Table~\ref{tab:main_results} presents the main experimental results across the three stages of the full fact-checking pipeline, as well as the overall performance determined jointly by the votes of judge agents in the evaluator panel.
From the results, we observe that:
1) GPT-o3 and DeepSeek-R1 achieve the best overall performance, ranking first and second, respectively, in terms of Elo scores, consistently outperforming other models across all stages.
2) The target LLMs generally show consistent performance across the three stages of the full fact-checking pipeline, except for Llama 4 Maverick and Gemini 2.5 Flash, whose results exhibit slight instability in the claim extraction stage. This indicates that despite variations in reasoning paths, these models demonstrate a degree of robustness, eventually converging into reasonable verifications through the complete reasoning process.
3) The stage-wise benchmarking for fact-checking can reveal distinct strengths and weaknesses among the models. For example, Gemini 2.5 Pro is better than GPT-4.5 in both claim extraction and claim verification, but weaker in evidence retrieval.
4) The accuracy metric reflects the performance as captured by traditional benchmarking methods. However, our results beyond the accuracy-only performance show that relying solely on such a metric provides an incomplete view of fact-checking capability. In contrast, our proposed benchmarking framework offers a more diverse and nuanced view by evaluating LLMs across open-form stages of the classical fact-checking pipeline, thereby enabling deeper scrutiny of their reasoning processes and robustness.
5) It can be further reflected in model-specific behavior. For example, although Gemini 2.5 Pro achieves the highest accuracy among all models, its advantage diminishes when evaluated across the full fact-checking pipeline. Similar inconsistencies are observed for Llama 4 Maverick and GPT-4.1, suggesting that accuracy on claim verification alone is insufficient to reflect a model’s overall fact-checking capability. Our proposed comprehensive and trustworthy auditing of the entire fact-checking workflow provides a more informative benchmarking perspective to supplement the evaluation, revealing strengths and weaknesses that static verification accuracy fails to capture.

\begin{figure*}[t!]
    \centering    \includegraphics[width=\textwidth]{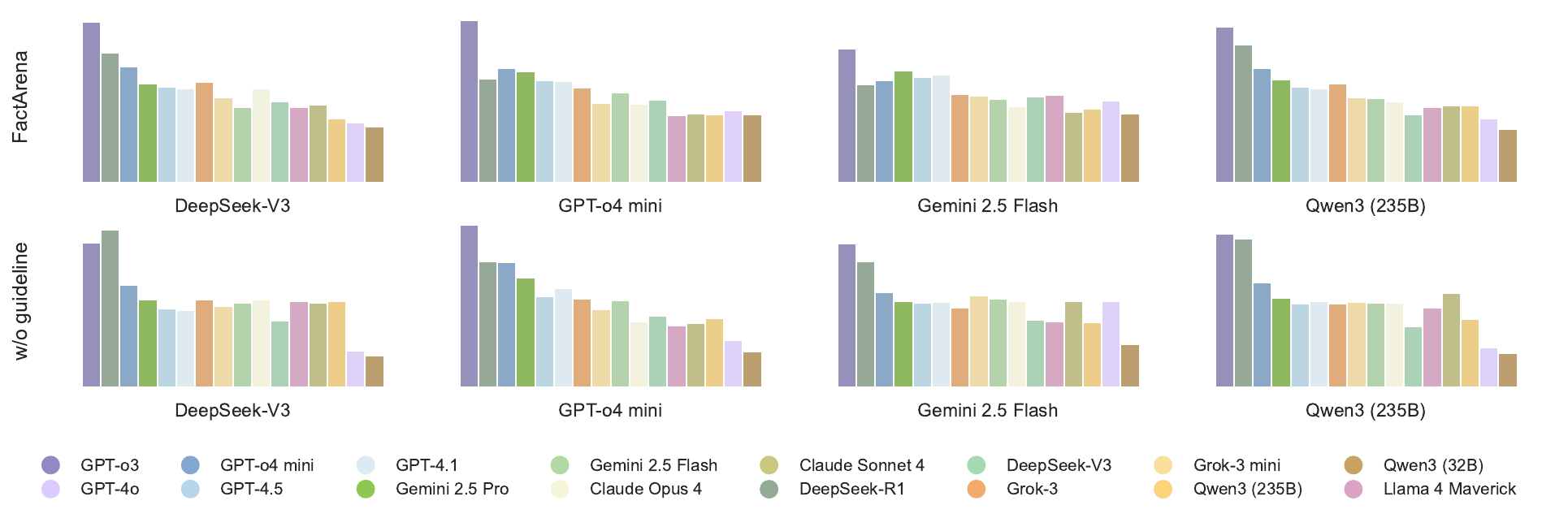} 
    \vspace{-0.7cm}
    \caption{An illustration of the Elo rankings under \textit{FactArena} and \textit{w/o guideline} settings. The order of target mLLMs is the ranking of our main result in Table~\ref{tab:main_results}.}
    \label{fig:rakings1}
    \vspace{-0.3cm}
\end{figure*}

\begin{table*}[t!]
\centering
\caption{The quantitative result of the judges in \textit{FactArena} and \textit{w/o guideline} settings. The inter-judge consistency scores are indicated by the accuracy(\%) of different judge results compared to majority vote results of the judge committee.}
\label{tab:abla_results}
\vspace{-0.3cm}
\renewcommand{\arraystretch}{0.9}
\resizebox{\textwidth}{!}{
\begin{tabular}{lccccc|ccccc}
\toprule
 & \multicolumn{5}{c|}{\textit{FactArena}} & \multicolumn{5}{c}{\textit{w/o guideline}} \\ \cmidrule(lr){2-6} \cmidrule(lr){7-11}
 & \textbf{DeepSeek} & \textbf{GPT-o4} & \textbf{Gemini 2.5} & \textbf{Qwen3} & {} & \textbf{DeepSeek} & \textbf{GPT-o4} & \textbf{Gemini 2.5} & \textbf{Qwen3} & {} \\
 & \textbf{-V3} & \textbf{mini} & \textbf{Flash} & \textbf{(235B)} & \textit{Avg.} & \textbf{-V3} & \textbf{mini} & \textbf{Flash} & \textbf{(235B)} & \textit{Avg.} \\ \cmidrule(lr){1-2} \cmidrule(lr){2-6} \cmidrule(lr){7-11}
\multicolumn{1}{l|}{\textit{Claim Extraction}} & 92.97 & 94.54 & 91.14 & 91.92 & 92.64 & 93.33 & 89.42 & 81.96 & 93.18 & 89.47 \\
\multicolumn{1}{l|}{\textit{Evidence Retrieval}} & 93.23 & 91.65 & 83.86 & 94.08 & 90.71 & 94.00 & 91.81 & 85.60 & 92.13 & 90.88 \\
\multicolumn{1}{l|}{\textit{Justification Production}} & 94.12 & 92.12 & 85.45 & 95.47 & 91.78 & 94.06 & 92.23 & 86.27 & 91.56 & 91.03 \\
\multicolumn{1}{l|}{\textit{Overall Pipeline}} & 93.60 & 93.66 & 87.03 & 95.73 & 92.51 & 95.25 & 92.36 & 87.44 & 92.87 & 91.98 \\ \bottomrule
\end{tabular}}
\end{table*}

\subsection{Analysis of Judgment Reliability with Guideline}


To qualitatively analyze the judgment reliability, we first conduct the ablative study to compare the \textit{FactArena}, and its \textit{`w/o guideline'} setting where the judge agents directly compare the analyses of target LLMs without using any reference guidelines.

Figure~\ref{fig:rakings1} illustrates the Elo rankings under \textit{FactArena} and \textit{w/o guideline} settings. We can observe that: 1) Overall, FactArena yields a more robust and well-stratified ranking structure, indicating that the use of reference guidelines helps unify judge decisions and produces more unbiased assessment outcomes. 2) Gemini 2.5 Flash deviates substantially from other judges, especially when evaluating lower-ranked models, suggesting systematic differences in judgment criteria. 3) Furthermore, Gemini 2.5 Flash and DeepSeek-V3 display notable preference patterns under the \textit{w/o guideline} setting, tending to assign higher rankings to models from the same family. This bias diminishes when guidelines are applied, reinforcing the necessity of standardized evaluation criteria for fair cross-model comparisons.

To further quantify inter-judge consistency for the analysis of the judgment reliability, we compute the accuracy of each judge’s decision relative to the majority-vote outcome, and report the average accuracy across judge agents. A higher average accuracy indicates stronger agreement among judges, and thus less idiosyncratic bias among LLM judges in evaluation.

As shown in Table~\ref{tab:abla_results}, incorporating our evaluation guidelines generally leads to more aligned judgments across different fact-checking stages. In particular, the claim extraction stage and the justification production stage exhibit notable improvements in average accuracy when guidelines are applied, suggesting that guidelines effectively align the judge committee in assessing model responses of different fact-checking stages. Although the average accuracy for the judgment in the evidence retrieval stage remains relatively unchanged when with the reference guideline, the designed reference guidelines generally help reduce individual subjective bias in evaluators' decisions across the overall pipeline. These results demonstrate that the proposed guidelines contribute to more stable and coherent judgments, reinforcing the reliability of the evaluation framework.


To further assess the reliability of our guideline-based evaluation, we conduct a human subject study in which human annotators replace the judge agents. Five human experts (aged 24–28) independently annotate a randomly sampled set of 100 battle pairs, following the same FactArena evaluation protocol as the judge agents.

Specifically, as shown in Table~\ref{tab:human_results}, the annotators, with an inter-annotator agreement of 0.632, and an intra-annotator agreement of 0.726, need to evaluate the fact-checking process as the judge agents.  The consistency is indicated by the accuracy of the judge agents using human annotations as the gold standard. We compare the \textit{FactArena} with the following two settings: 1) `\textit{w/o guideline}': The judge agents directly compare the analyses of target mLLMs without using any references; 2) `\textit{LLM-as-a-judge}': The judge agents use the analyses generated by themselves as references for judgments. We can see that the joint voting of the `\textit{FactArena'} judge committee has the best consistency with the human annotations. The inter-judge consistency under the three ablative settings in Table~\ref{tab:abla_results2} also validates the advantage of our proposed FactArena.
The resulting alignment indicates that the guideline-driven judging procedure provides more consistent and human-aligned evaluations of the fact-checking workflow.



\begin{table}[]
\caption{The consistency between judge agents and human evaluators under different settings.}
\label{tab:human_results}
\vspace{-0.3cm}
\resizebox{\linewidth}{!}{
\begin{tabular}{l|ccccc} \toprule
 & \textbf{DeepSeek} & \textbf{GPT-o4} & \textbf{Gemini 2.5} & \textbf{Qwen3} & \textit{Joint} \\
 & \textbf{-V3} & \textbf{mini} & \textbf{Flash} & \textbf{(235B)} & \textit{Voting} \\ \hline
\textit{FactArena} & 0.67 & 0.85 & 0.72 & 0.69 & 0.75 \\
\textit{w/o guideline} & 0.74 & 0.74 & 0.67 & 0.66 & 0.69 \\
\textit{LLM-as-a-judge} & 0.70 & 0.69 & 0.69 & 0.63 & 0.66\\ \bottomrule
\end{tabular}}
\end{table}

\begin{table}[]
\caption{The inter-judge consistency(\%) under different settings.}
\label{tab:abla_results2}
\vspace{-0.3cm}
\resizebox{\linewidth}{!}{
\begin{tabular}{l|ccccc} \toprule
 & \textbf{DeepSeek} & \textbf{GPT-o4} & \textbf{Gemini 2.5} & \textbf{Qwen3} & { } \\
 & \textbf{-V3} & \textbf{mini} & \textbf{Flash} & \textbf{(235B)} & \textit{Avg.} \\ \hline
\textit{FactArena} & 93.60 & 93.66 & 87.03 & 95.73 & 92.51 \\
\textit{w/o guideline} & 95.25 & 92.36 & 87.44 & 92.87 & 91.98 \\ 
\textit{LLM-as-a-judge} & 88.35 & 95.61 & 91.88 & 93.73 & 92.39\\ \bottomrule
\end{tabular}}
\end{table}


\subsection{Analysis of Arena-driven Claim Evolution}


To verify the effectiveness of the arena-driven claim evolution, we conduct ablative studies in the following settings: a) `\textit{FactArena}': the benchmark framework with arena-driven claim evolution; b) `\textit{w/o evolution}': simply reverse the trivial claim whose factuality is correctly predicted by all target models, but without further data evolution; c) `\textit{w/o reverse}': directly removing the first reverse step and no further evolution step.

\begin{figure}[t!]
    \centering    \includegraphics[width=\linewidth]{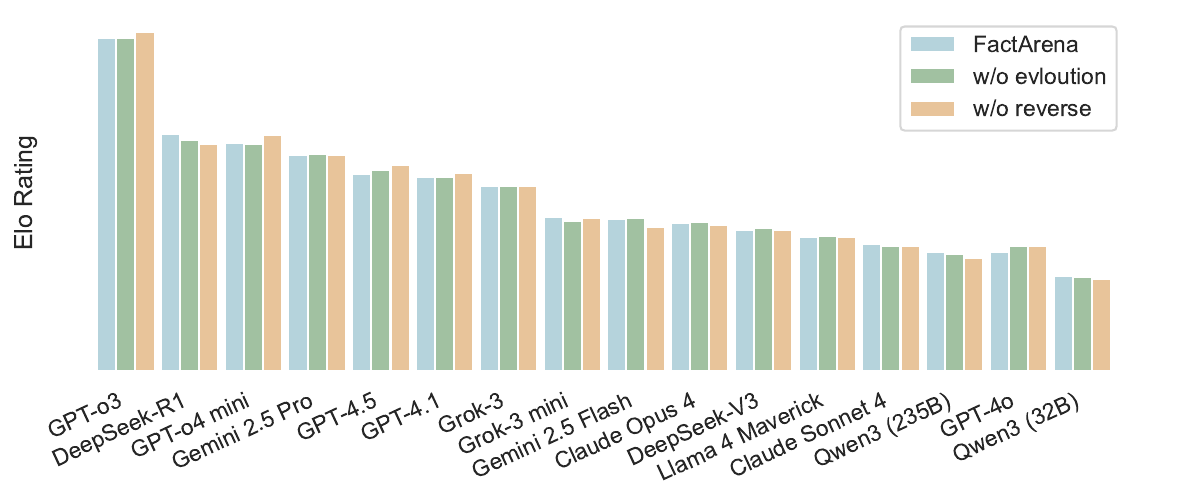} 
    \vspace{-0.7cm}
    \caption{Elo ratings of target LLMs before and after claim evolution.}
    \label{fig:rakings_abla}
    \vspace{-0.3cm}
\end{figure}

Figure~\ref{fig:rakings_abla} provides an intuitive illustration of the Elo ratings of target models under our ablation settings. We observe that the ratings of DeepSeek-R1, Gemini 2.5 Flash, Qwen 3 (235B), and Qwen 3 (32B) increase noticeably after claim evolution, indicating that these models are more robust and generalizable when confronted with adaptively generated challenging claims. Conversely, models whose ratings drop under evolution exhibit stronger vulnerability. In contrast, a decrease under the `\textit{FactArena}' setting may also suggest that there could be potential data leakage under the `\textit{w/o evolution}' and `\textit{w/o reverse}' settings, models perform well on static benchmarks but fail once claims are transformed into more novel and unseen forms. Since our claim-evolution procedure continuously adapts claims toward harder and out-of-distribution variations, models that sustain or improve their scores under this setting are likely to possess more robust and broadly generalizable fact-checking capabilities.

\begin{figure}[t!]
    \centering    \includegraphics[width=\linewidth]{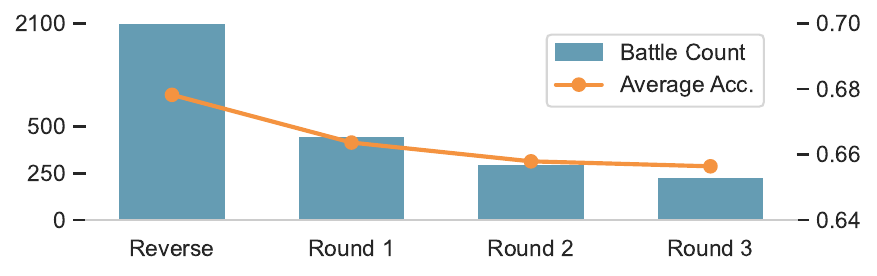} 
    \vspace{-0.7cm}
    \caption{Effect of arena-driven claim evolution rounds.}
    \label{fig:evol_trend}
    \vspace{-0.3cm}
\end{figure}

We further analyze the effect of the arena-driven claim evolution rounds, as illustrated in Figure~\ref{fig:evol_trend}, where the Reverse means the Round 0 of the claim evolution. It can be observed that simple claims, those correctly predicted by all target models (100\% accuracy), experience a substantial drop in average prediction accuracy to 68\% after semantic reversal, indicating that the reversed claims become significantly more challenging. As the evolution rounds progress, the average accuracy continues to decrease, though the rate of decline gradually slows. The battle-count statistics further corroborate this trend: the number of simple claims decreases steadily across rounds and the battle count stabilizes at around 220 by rounds 2–3, suggesting that the evolution process converges as fewer claims remain trivially solvable by all the target models.

\subsection{Qualitative Comparison with Traditional Benchmark}
To guarantee the reliability of the benchmarking results from FactArena, we design a blind test that compares FactArena against traditional benchmarking like \cite{lin2025fact} that focuses only on the claim verification stage. We randomly sample 100 battles, and five human experts are asked to select the better judgment of the same judge agent under the two evaluation designs based on the following criteria:
1) \textit{Helpfulness}: whether the judge’s explanation about the battle results aligns with the true veracity label of the claim. Human experts are asked to select the response that is more convincing and less misleading;
2) \textit{Informativeness}: whether the judge's explanation provides additional useful information, such as background details or contextual knowledge;
3) \textit{Soundness}: whether the judge's explanation appears valid, well-reasoned, and logically coherent;
4) \textit{Readability}: whether the judge's explanation follows proper grammar and structure, and whether the sentences are clear and easy to follow.

\begin{figure}[t!]
    \centering    \includegraphics[width=\linewidth]{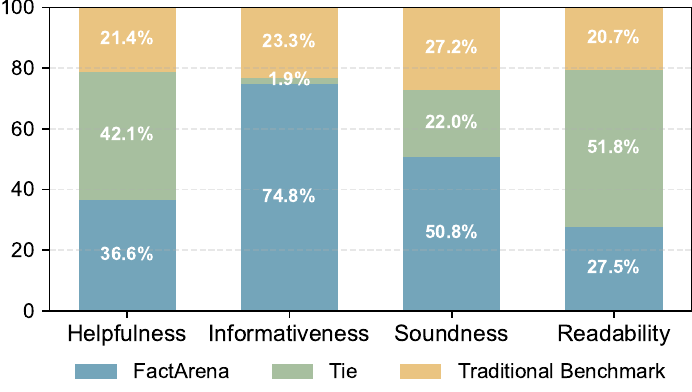} 
    \vspace{-0.7cm}
    \caption{Human preference of judgments under FactArena and the traditional benchmarking paradigm.}
    \vspace{-0.3cm}
    \label{fig:vs_traditional}
\end{figure}

Note that here we use the same guidelines for the human experts as those for the judge agents in the claim verification stage, but the difference lies in that here we need to evaluate the judge agents themselves. As shown in Figure~\ref{fig:vs_traditional}, judge responses produced under the FactArena framework exhibit a significant advantage in \textit{Informativeness}, indicating that full-pipeline fact-checking naturally incorporates richer information and enables a more comprehensive assessment of model behavior. Moreover, responses from FactArena judges are not only more sound but also more convincing and less misleading compared to those from the traditional benchmarking paradigm.

\subsection{Discussion of Judge Model Version}

\begin{table}[t] \tiny
\caption{The consistency between judges and their stronger versions from the same model family.}
\label{tab:stronger_model_comparison}
\vspace{-0.3cm}
\centering
\resizebox{\linewidth}{!}{
\renewcommand{\arraystretch}{1.1}
\begin{tabular}{l@{\hspace{20pt}}c} \toprule
\textit{Model Pair} & \textit{Consistency} \\  \hline
\textbf{DeepSeek-V3 $\leftrightarrow$ DeepSeek-R1} & 0.92 \\
\textbf{GPT-o4 mini $\leftrightarrow$ GPT-o3} & 0.74 \\
\textbf{Gemini 2.5 Flash $\leftrightarrow$ Gemini 2.5 Pro} & 0.72 \\ \hline
{\textit{Avg.}} & 0.79 \\ \bottomrule
\end{tabular}}
\end{table}


In deploying the judge agents for FactArena, we deliberately selected models that balance evaluation quality and computational cost, rather than exclusively relying on the strongest available LLMs. To assess whether this choice affects evaluation reliability, we compare each judge agent against a stronger model from the same family. Table~\ref{tab:stronger_model_comparison} presents the consistency between our chosen judges and their more capable counterparts. The results show that the judgments produced by the selected judge agents remain highly consistent with those of stronger models, with average agreement approaching 80\%, and in some cases, such as DeepSeek-V3, exceeding 90\%.

These findings indicate that the evaluation quality is not significantly compromised by using computationally lighter judge agents, and that their decisions remain well aligned with both human assessments and stronger LLMs within the same model family.

\subsection{Discussion of Judge Agent Numbers}

To further explore the effect of the judge agents in the evaluator panel, we conduct evaluations on the number of judges. As shown in Table~\ref{tab:num_judges}, we compare the consistency of joint agent judgments with human evaluators. The settings are as follows: 1) 1 Judge: Deepseek-V3; 2) 2 Judges: Deepseek-V3 and GPT-4o; 3) 3 Judges: Deepseek-V3, GPT-4o and Gemini 2.5 Flash; 4) 4 Judges: the 4 judge agents in our main experiment.

Compared with configurations using fewer judges, the 4-judge setting in FactArena achieves the highest accuracy. We also observe that Deepseek-V3 alone reaches an accuracy of 0.67 as a single judge, but when GPT-4o is added to the evaluator panel, the combined accuracy slightly decreases, indicating divergence in judge opinions. However, as the number of judges continues to grow, the panel’s overall accuracy increases, suggesting that aggregating more judge agents leads to decisions that better reflect human preferences. While involving a larger set of judges improves alignment, greater computational and resource costs are inevitable. From a cost effective standpoint, and consistent with the principle of Occam’s razor, the 4-judge configuration offers a strong balance between evaluation quality and efficiency, making it a practical choice.

\begin{table}[]
\caption{The effect of different numbers of judge agents.}
\label{tab:num_judges}
\centering
\renewcommand{\arraystretch}{1.2}
\vspace{-0.3cm}
\setlength{\tabcolsep}{11pt}
\resizebox{\linewidth}{!}{\begin{tabular}{l|cccc}
\toprule
 & \textbf{1 Judge} & \textbf{2 Judges} & \textbf{3 Judges} & \textbf{4 Judges}  \\
\midrule
\textit{Accuracy} & 0.67 & 0.65 & 0.69 & 0.75 \\
\bottomrule
\end{tabular}}
\end{table}

\subsection{Analysis of Claim Extraction}

To examine how the quality of the claim extraction guideline evolves across different rounds of integration, we conduct a detailed human evaluation. We recruit five human experts aged 24–28 and randomly sample 20 claims for assessment at each round (the average intra-annotator agreement is 0.618). 

\begin{figure}[t!]
    \centering    \includegraphics[width=\linewidth]{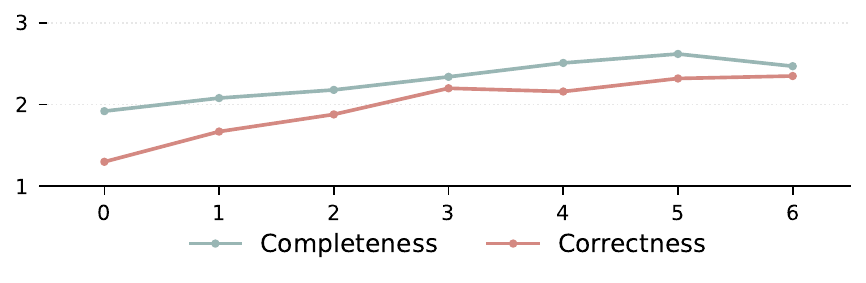} 
    \vspace{-0.7cm}
    \caption{The \textit{Completeness} and \textit{Correctness} scores of the evaluation guideline for claim extraction during multiple rounds iterations.}
    \label{fig:fusion_trend}
    \vspace{-0.3cm}
\end{figure}

Following \cite{kamoi2023wicerealworldentailmentclaims}, we evaluate guideline quality using two task-decomposition metrics:
(1) \textit{Completeness}: whether the sub-claims collectively cover all essential information in the original claim;
(2) \textit{Correctness}: whether each sub-claim faithfully represents a specific part of the original claim without distortion or unnecessary additions.
As shown in Figure~\ref{fig:fusion_trend}, we can see that the results show a clear trend of an increase in guideline quality from rounds 0 to 3. After the third round, the improvement rate slows, and by rounds 5–6, the scores converge, indicating that additional refinement brings diminishing returns.


\subsection{Case Study}
\begin{figure*}[t!]
    \centering    \includegraphics[width=\textwidth]{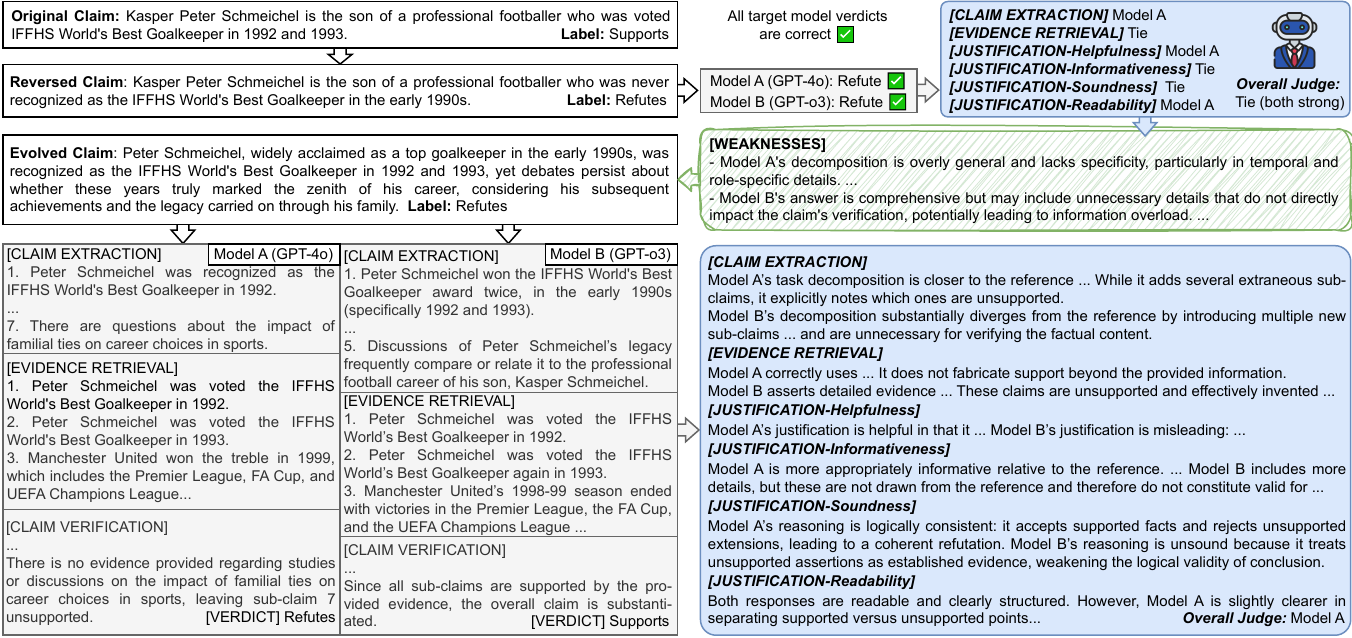} 
    \vspace{-0.7cm}
    \caption{An example of arena-driven claim evolution. Both Model A and Model B correctly predict the original and reversed claims. After model-specific weaknesses are analyzed and a more challenging evolved claim is derived, clear performance distinctions emerge between the target models.}
    \label{fig:case_study}
    \vspace{-0.3cm}
\end{figure*}

To probe deeper into the knowledge boundaries of target models, FactArena employs an arena-driven claim evolution procedure that progressively transforms simple claims into more challenging ones.
As shown in Figure~\ref{fig:case_study}, the original claim is a straightforward fact-verification question that is correctly answered by all tested target models. After reversing the claim's factuality, both Model A and Model B still respond correctly. The judge agent, therefore, considers both models to be strong and makes a tie judgment. However, during the analysis of model weaknesses, the evolver agent observes that Model A’s reasoning may lack structured reasoning, while Model B tends to include unnecessary details and might potentially leading to information overload. Motivated by these observations, the evolved claim rewrites and expands the original statement, reformulating the claim from a simple knowledge verification problem into a more logically complex statement.
When evaluated on the evolved claim, Model A correctly identifies that the seventh sub-claim is unsupported, whereas Model B’s analysis contains substantial redundant information and misses the false relation in the claim. This example demonstrates that in arena-driven evolution, the claim difficulty is selectively intensified, which effectively amplifies performance distinctions between models and thus enables a more precise assessment of target models’ true understanding of factual knowledge.

\section{Conclusion and Future Work}
In this work, we introduced FactArena, a fully automated, arena-style benchmarking framework that systematically evaluates LLMs across the entire fact-checking pipeline. By integrating stage-wise auditing, guideline-driven judgment, and arena-based claim evolution, FactArena provides a more comprehensive and trustworthy assessment of LLMs’ factual reasoning capabilities than traditional verification-only evaluation. Extensive experiments over 16 state-of-the-art models show that FactArena produces robust and fair rankings, substantially improves inter-judge consistency, and reveals performance discrepancies that static accuracy metrics fail to capture. Moreover, the adaptive claim-evolution mechanism effectively probes model robustness and uncovers deeper reasoning limitations that remain hidden under fixed test sets.

Future work can further extend FactArena along several directions. First, although our framework focuses on textual fact-checking, incorporating multimodal claims~\cite{shao2024detecting, wang2025mfc} and richer retrieval sources (e.g., structured databases, temporal information) may broaden its applicability. Second, enhancing judge-agent diversity and integrating uncertainty-aware evaluation may provide finer-grained diagnostic signals for model reliability. Finally, deploying FactArena in real-world fact-checking scenarios, where complex claims are noisier, adversarial, or rapidly evolving, opens opportunities for studying model behavior under more realistic and dynamic conditions. We hope this framework serves as a foundation for building rigorous, scalable, and human-aligned evaluations of future LLMs in safety-critical factual reasoning tasks.

\bibliography{sample}
\bibliographystyle{IEEEtran}

\clearpage

\appendices

\section{Prompt Templates}

We provide the curated prompt templates in our FactArena framework here.

\begin{figure}[t!]
    \centering    \includegraphics[width=\linewidth]{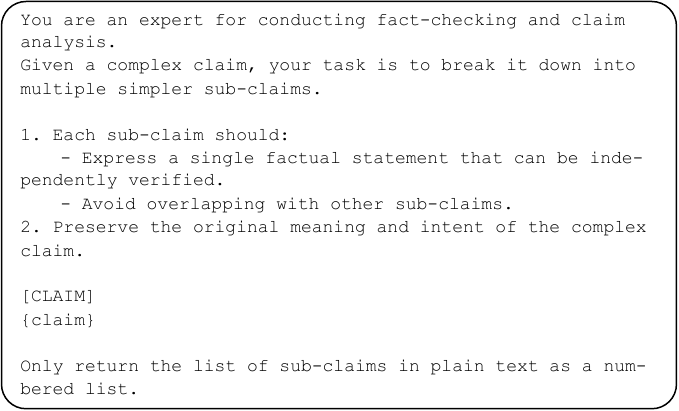} 
    \caption{The prompt for conducting claim extraction.}
    \label{fig:prompt_claim_extraction}
    \vspace{-0.3cm}
\end{figure}

\paragraph{Claim Extraction Prompt}
The prompt used for claim extraction is provided in Figure~\ref{fig:prompt_claim_extraction}. The target models are instructed to break down the claims into sub-claims of proper granularity.

\begin{figure}[t!]
    \centering    \includegraphics[width=\linewidth]{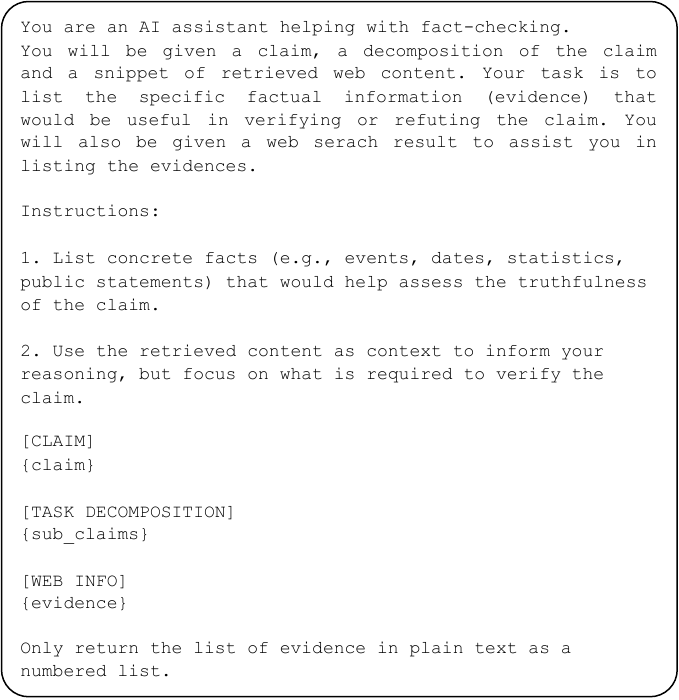} 
    \caption{The prompt for constructing evidence for claim verification.}
    \label{fig:prompt_evidence_retrieval}
    \vspace{-0.3cm}
\end{figure}

\paragraph{Evidence Retrieval Prompt}
The prompt used for evidence retrieval is provided in Figure~\ref{fig:prompt_evidence_retrieval}. In evidence retrieval, the target models generate a list of factual evidence based on the claim, the sub-claims, and the web information retrieved using the external tool.

\begin{figure}[t!]
    \centering    \includegraphics[width=\linewidth]{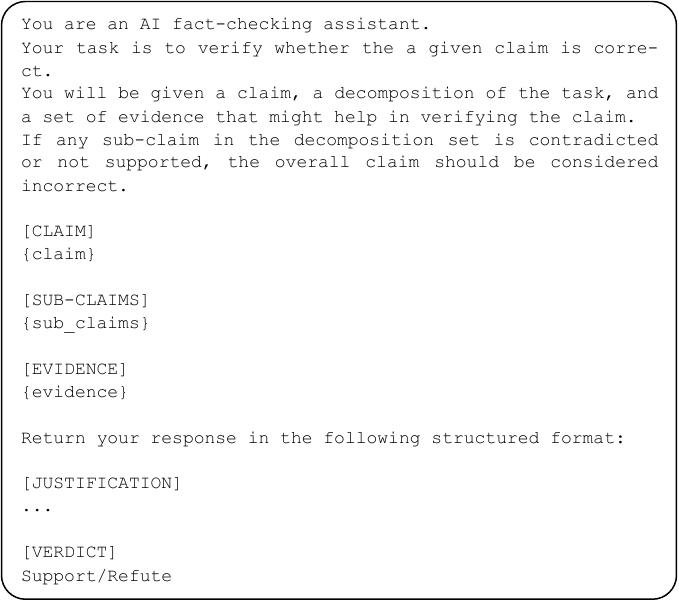} 
    \caption{The prompt for justification production and verdict prediction.}
    \label{fig:prompt_justification}
\end{figure}

\paragraph{Justification Production \& Verdict Prediction Prompt}
The prompt used to generate justification and verdict prediction is shown in Figure~\ref{fig:prompt_justification}. Based on the previously generated sub-claims and evidence, the target model is instructed to justify and give a final verdict of the claim.

\begin{figure}[t!]
    \centering    \includegraphics[width=\linewidth]{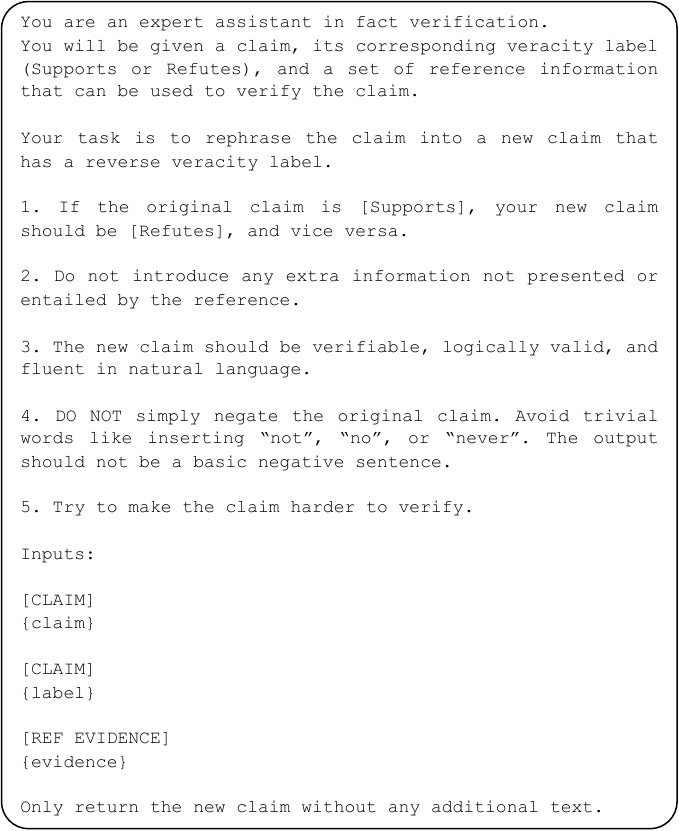} 
    \caption{The prompt for reversing the semantics of claims.}
    \label{fig:prompt_reverse}
    \vspace{-0.3cm}
\end{figure}

\begin{figure}[t!]
    \centering    \includegraphics[width=\linewidth]{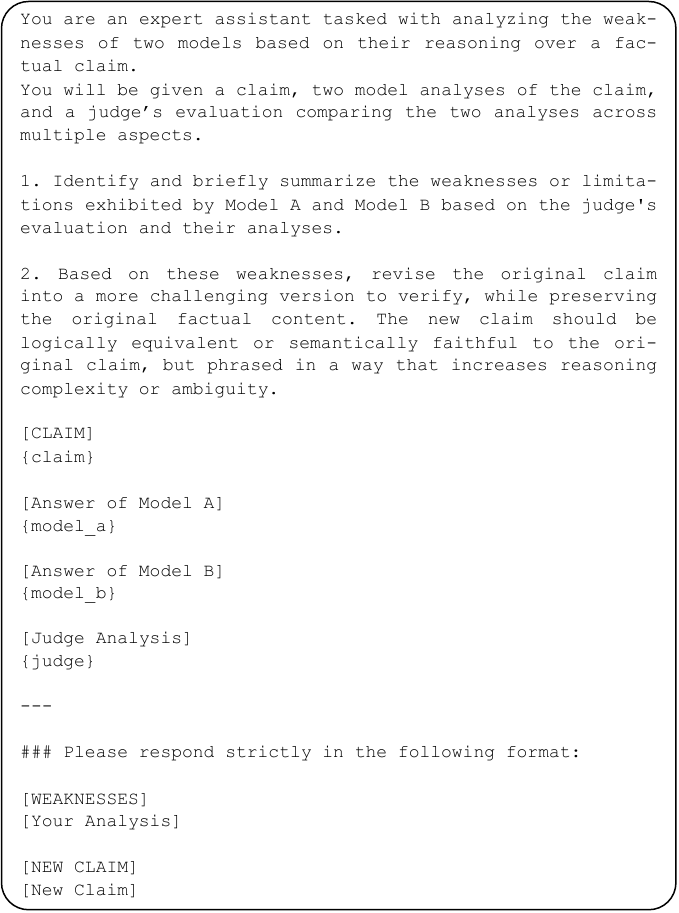} 
    \caption{The prompt for arena-driven claim evolution.}
    \label{fig:prompt_evol}
\end{figure}

\paragraph{Evolution Prompt}
Figure~\ref{fig:prompt_reverse} and Figure~\ref{fig:prompt_evol} show the prompts for reversing the semantics and generating more challenging claims. When reversing the claim’s semantics, the evolver agent reformulates the claim into a logical and verifiable alternative that has an opposite veracity label, avoiding making superficial changes or trivial modifications. In the arena-driven claim evolution process, the evolver agent analyzes the specific weaknesses of the target models and subsequently rephrases the claim into a more challenging version designed to probe these weaknesses.

\begin{figure}[t!]
    \centering    \includegraphics[width=\linewidth]{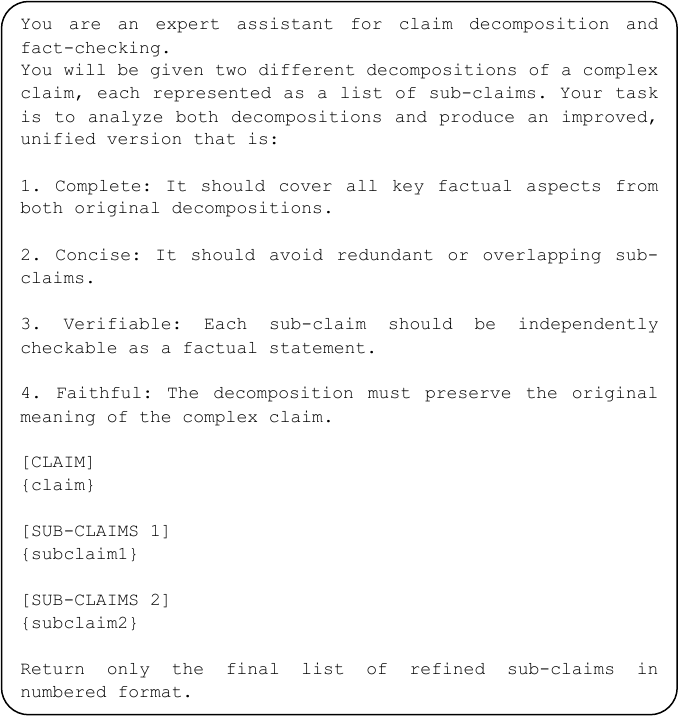} 
    \caption{The prompt for consolidating the sub-claims into the evaluation guideline for claim extraction.}
    \label{fig:prompt_fusion}
\end{figure}

\paragraph{Claim Extraction Guideline Consolidation Prompt}
The prompt for consolidating the claim extraction guideline is shown in Figure~\ref{fig:prompt_fusion}. To mitigate potential bias, the judge agents are blinded to the sources of the sub-claims, and their order of model answers is randomly shuffled.

\begin{figure}[t!]
    \centering    \includegraphics[width=\linewidth]{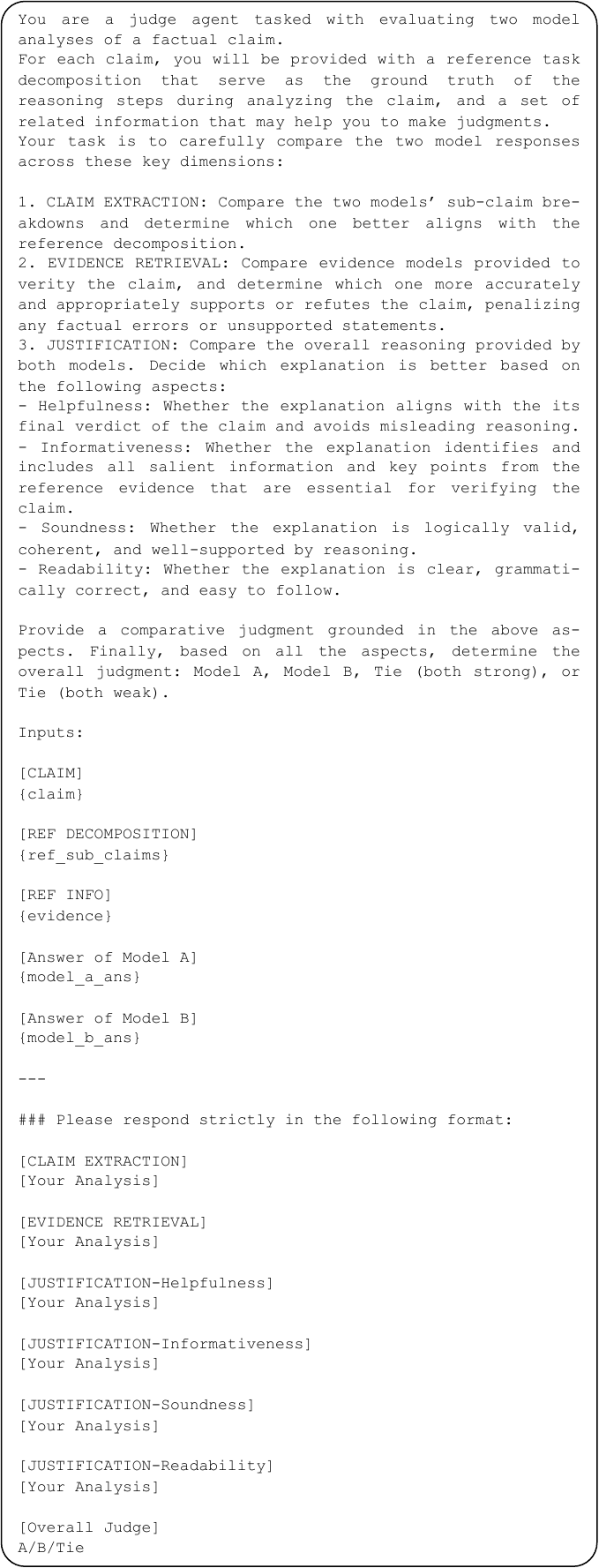} 
    \caption{The prompt for arena-styled stage-wise judgment.}
    \label{fig:prompt_judge}
\end{figure}

\paragraph{Arena-styled Stage-wise Judgment Prompt}
The prompt for agent judgment is provided in Figure~\ref{fig:prompt_judge}. To avoid potential positional bias, the order of model answers is randomly given. The judge gives its judgment by comparing the performance of the target models at each stage of claim verification.

\begin{figure}[t!]
    \centering  \vspace{0.3cm}
    \includegraphics[width=\linewidth]{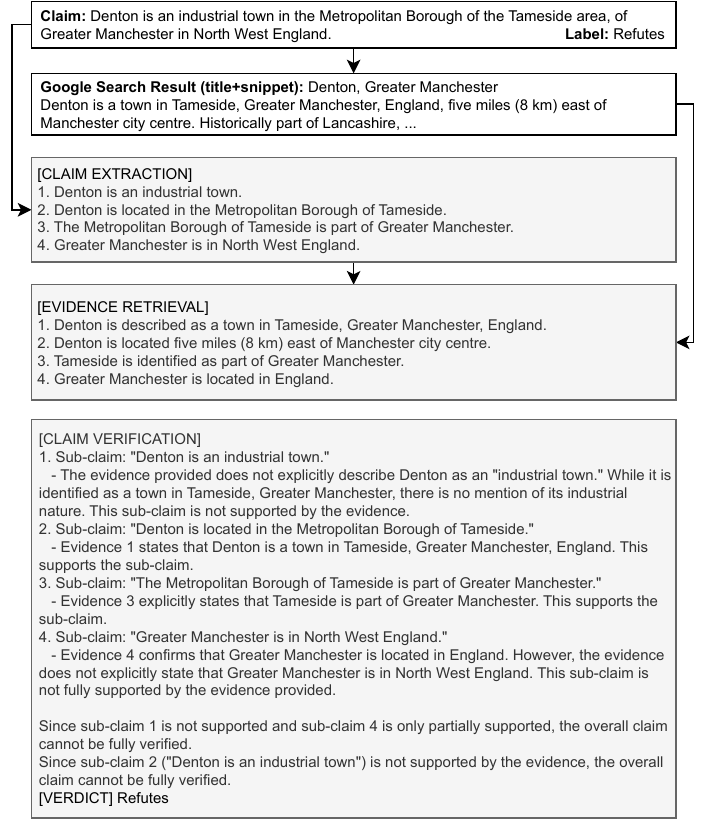}
    \caption{A case of FactArena fact-checking pipeline.}
    \label{fig:case_model_answering}
    \vspace{-0.3cm}
\end{figure}

\section{Detailed Cases}
In this section, we provide more case studies regarding each specific procedure in the FactArena framework.

\setcounter{paragraph}{0}
\paragraph{A Case of Fact-checking Pipeline}
The three-stage fact-checking pipeline of FactArena is illustrated in Figure~\ref{fig:case_model_answering}, consisting of claim extraction, evidence retrieval, and claim verification. In the claim extraction stage, the target model decomposes the original claim into a set of sub-claims. Based on these sub-claims, the model then retrieves relevant web information via external tools and organizes it into supporting evidence. Finally, during claim verification, the model integrates the original claim with the extracted sub-claims and retrieved evidence to produce a final veracity verdict.

\begin{figure*}[t!]
    \centering    \includegraphics[width=\textwidth]{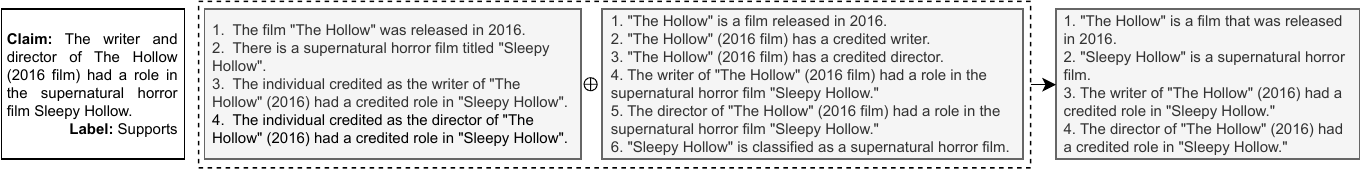} 
    \vspace{-0.7cm}
    \caption{A case of guideline consolidation in constructing the evaluation guideline of claim extraction.}
    \label{fig:case_fusion}
    \vspace{-0.3cm}
\end{figure*}

\paragraph{A Case of Guideline Consolidation}
Figure~\ref{fig:case_fusion} demonstrates an example of synthesizing the claim extraction evaluation guideline. As shown in the case, the refined guideline preserves the decomposition strategy of the current reference guideline, while incorporating the expression style in the model answers. The synthesized decomposition results in a more coherent and practically verifiable reference guideline for claim extraction.

\begin{figure*}[t!]
    \centering    \includegraphics[width=\textwidth]{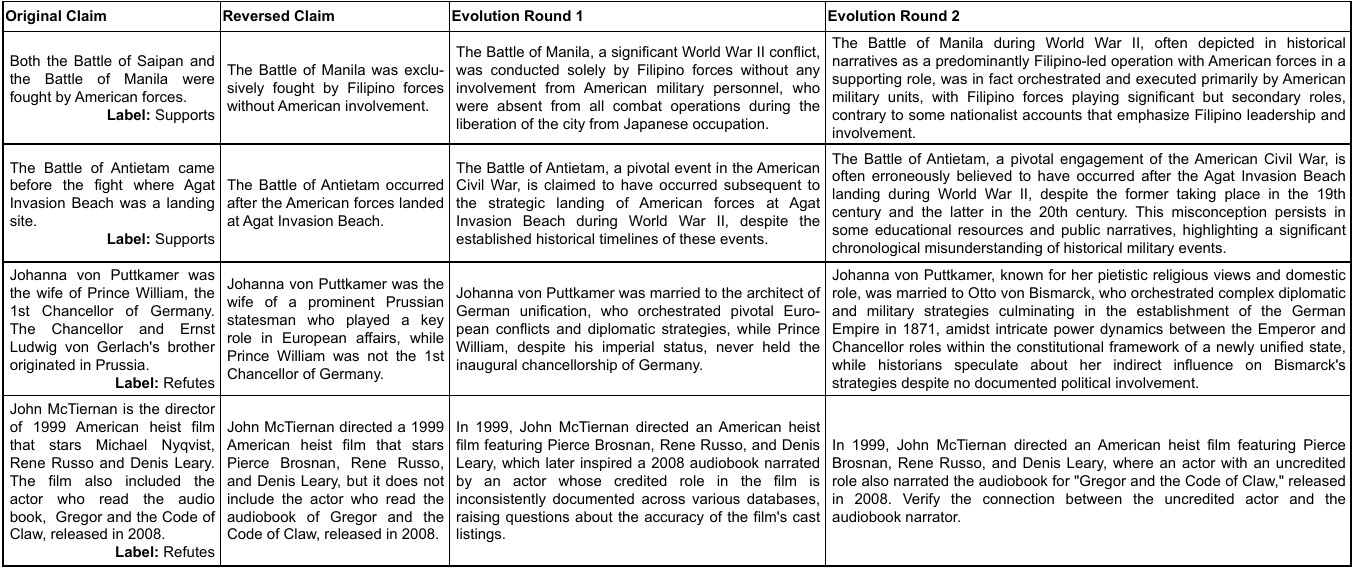} 
    \vspace{-0.7cm}
    \caption{More cases of claim evolution.}
    \label{fig:more_cases_evol}
    \vspace{-0.3cm}
\end{figure*}

\paragraph{More Cases of Claim Evolution}
We further present additional cases of claim evolution in Figure~\ref{fig:more_cases_evol}, illustrating the original claim, the semantically reversed claim, and the claims in two rounds of arena-driven evolution. As the evolution progresses, the claim semantics become increasingly complex, making the verification task more challenging.

\begin{figure*}[t!]
    \centering    \includegraphics[width=\textwidth]{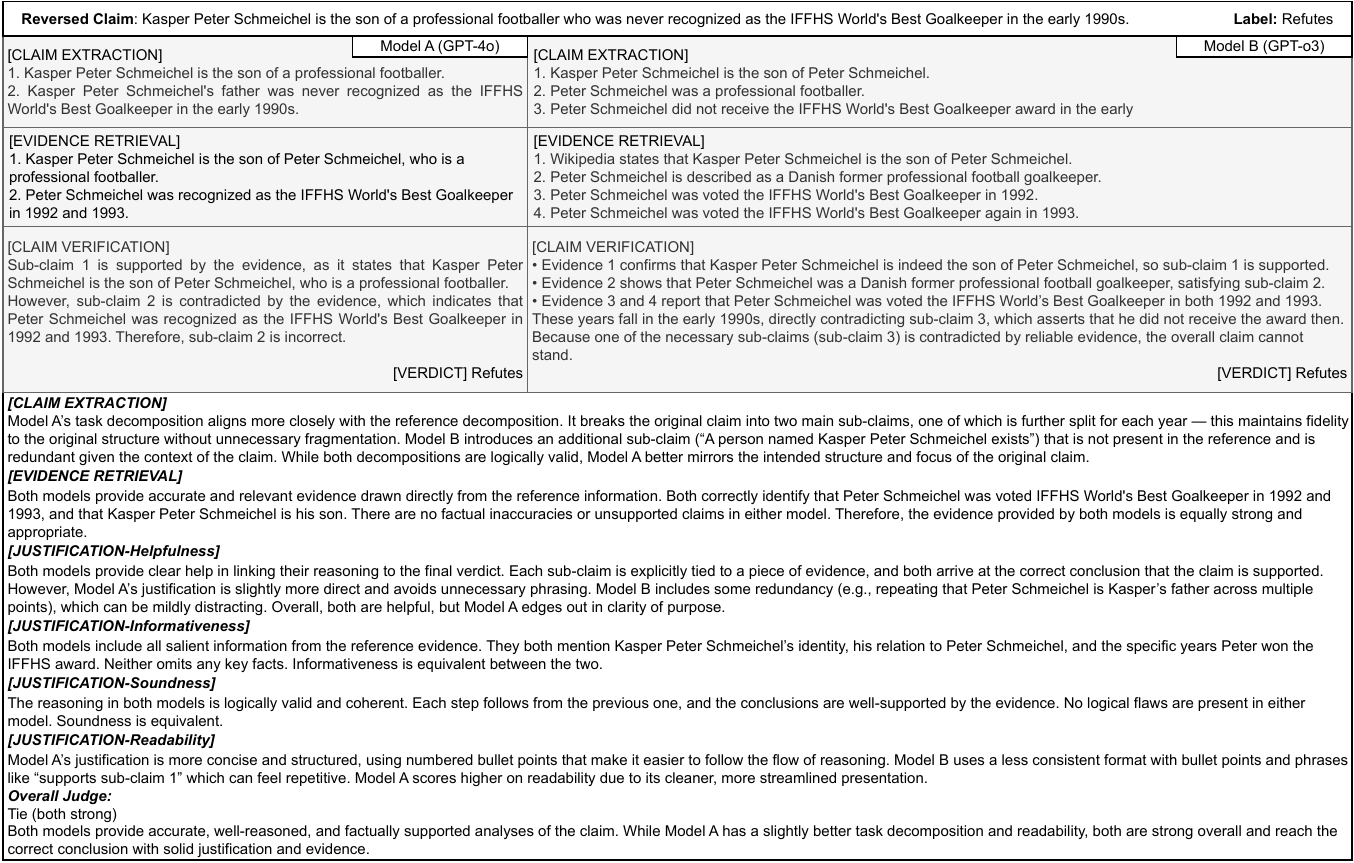} 
    \vspace{-0.7cm}
    \caption{A detailed example of arena-driven claim evolution in reversing the claim's semantics.}
    \label{fig:evol_round0}
\end{figure*}

\begin{figure*}[t!]
    \centering    \includegraphics[width=\textwidth]{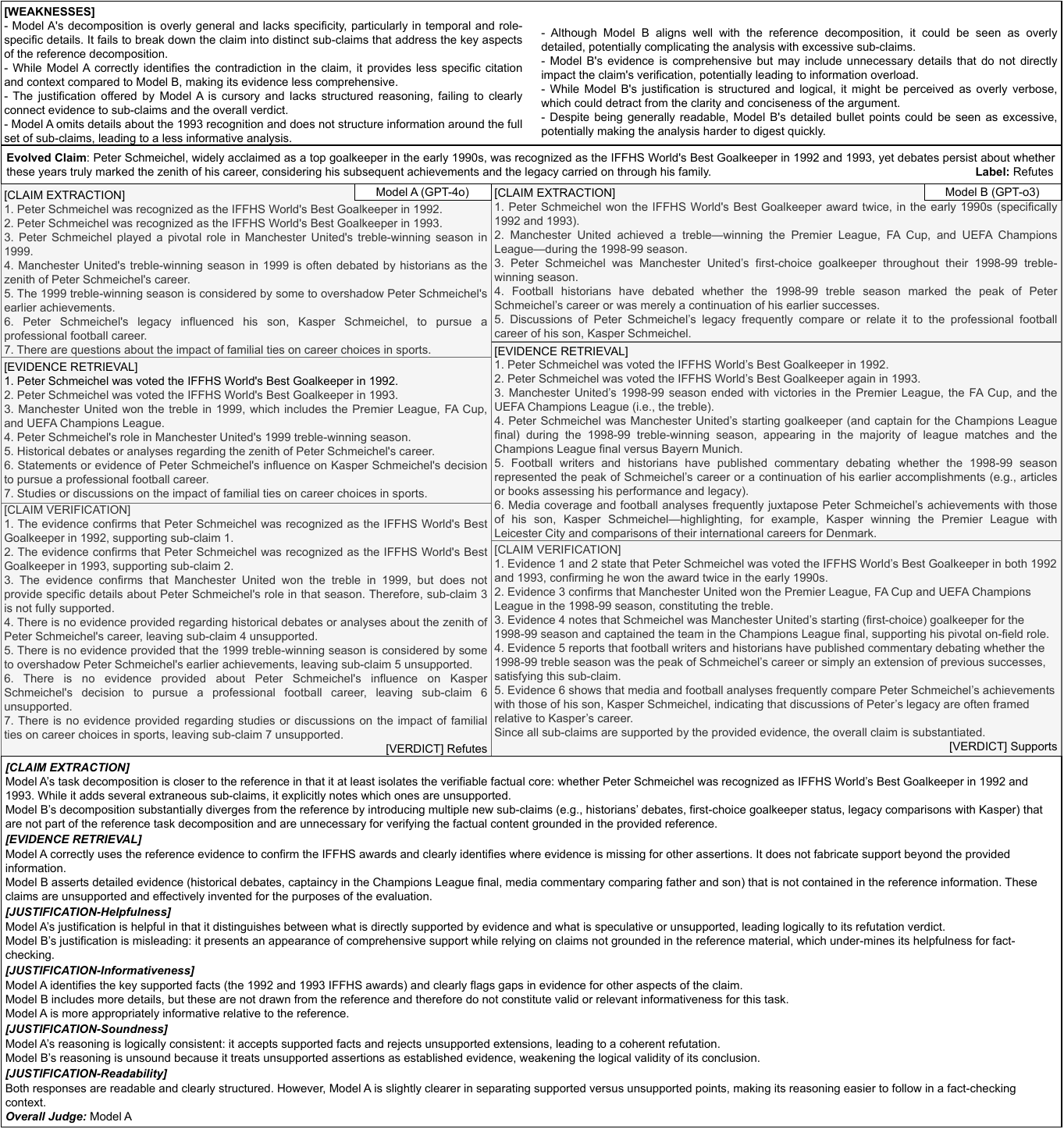} 
    \caption{A detailed example of arena-driven claim evolution in generating a more challenging claim.}
    \label{fig:evol_round1}
\end{figure*}

\paragraph{Detailed Case of Arena-driven Claim Evolution}
We show the more detailed case of the arena-driven claim evolution in our benchmarking process in Figure~\ref{fig:evol_round0} and Figure~\ref{fig:evol_round1}.

\vfill

\end{document}